\documentclass[letterpaper, 10 pt, conference]{ieeeconf}  %
\IEEEoverridecommandlockouts   
\title{\LARGE \bf
Vision-Conditioned Variational Bayesian Last Layer Dynamics Models
}

\author{
Paul Brunzema$^1$, Thomas Lew$^2$, Ray Zhang$^2$, Takeru Shirasawa$^2$, John Subosits$^2$, Marcus Greiff$^{2,*}$
\thanks{$^{*}$ Corresponding author.}
\thanks{$^{1}$Paul Brunzema is with the department of Mechanical Engineering, Aachen University, Germany. Email: 
         {\small paul.brunzema@googlemail.com}. Paul was with TRI when conducting this research.}%
 \thanks{$^{2}$The remaining authors are with Toyota Research Institute,
         4440 El Camino Real, CA, USA. Email:
         {\small first.last@tri.global}.}%
}

\usepackage{mathtools}
\usepackage{siunitx}
\usepackage{lipsum}
\usepackage{wrapfig}
\usepackage{cite}

\let\labelindent\relax
\usepackage{enumitem}

\usepackage{transparent}
\usepackage[utf8]{inputenc}
\usepackage[T1]{fontenc}
\usepackage{hyperref}
\usepackage{url}
\usepackage{booktabs}
\usepackage{amsfonts}
\usepackage{nicefrac}
\usepackage{microtype}
\usepackage{graphicx}
\usepackage{doi}
\usepackage{amsmath}
\usepackage{amssymb}
\usepackage{accents}
\usepackage{xcolor}
\usepackage{xspace}
\usepackage{mdframed}
\usepackage{tikz}
\usepackage{multirow}
\usepackage{threeparttable, tablefootnote}
\usepackage{algpseudocode,algorithm,algorithmicx}
\usepackage{varwidth}
\usepackage[normalem]{ulem}

\newcommand\segman{\textsc{Segman}\xspace}

\newcommand\goose{\textsc{Goose}\xspace}
\newcommand\segnext{\textsc{SegNeXt}\xspace}

\hypersetup{
     colorlinks   = true,
     citecolor    = blue,
     allcolors = blue,
}

\algrenewcommand\algorithmicrequire{\textbf{Input:}}
\algrenewcommand\algorithmicensure{\textbf{Output}}

\makeatletter
\algnewcommand{\AlgComment}[1]{\Statex \hskip\ALG@thistlm {\color{gray}\textit{// #1}}}
\algnewcommand{\AlgCommentIndent}[1]{\Statex \hskip\ALG@tlm {\color{gray}\textit{// #1}}}
\makeatother

\newcommand{\ubar}[1]{\underaccent{\bar}{#1}}
\newcommand\myvec[1]{\boldsymbol{#1}}

\newcommand\cvec{\myvec{c}}

\newcommand\fvec{\myvec{f}}

\newcommand\mvec{\myvec{m}}

\newcommand\uvec{\myvec{u}}
\newcommand\vvec{\myvec{v}}
\newcommand\wvec{\myvec{w}}
\newcommand\xvec{\myvec{x}}
\newcommand\yvec{\myvec{y}}
\newcommand\zvec{\myvec{z}}

\newcommand\phivec{\myvec{\varphi}}
\newcommand\gammavec{\myvec{\gamma}}
\newcommand\betavec{\myvec{\beta}}

\newcommand{\Bcal}{\mathcal{B}}
\newcommand{\Ccal}{\mathcal{C}}
\newcommand{\Dcal}{\mathcal{D}}

\newcommand{\Gcal}{\mathcal{G}}

\newcommand{\Scal}{\mathcal{S}}

\newcommand\I{\myvec{I}}

\newcommand\F{\myvec{F}}

\newcommand\Lbf{\myvec{L}}

\newcommand\Q{\myvec{Q}}
\newcommand\R{\myvec{R}}
\newcommand\W{\myvec{W}}
\newcommand\N{\myvec{N}}
\newcommand\Z{\myvec{0}}
\newcommand\T{\myvec{T}}

\newcommand\Sbf{\boldsymbol{S}}
\newcommand\Pbf{\boldsymbol{P}}

\newcommand\XX{\myvec{X}}

\newcommand\der{\mathrm{d}}

\newcommand\Gaussian{\mathrm{N}}

\newcommand\Real{\mathbb{R}}

\newcommand\states{\xvec}
\newcommand\controls{\uvec}
\newcommand\context{\cvec}
\newcommand\inputs{\zvec}
\newcommand\features{\phivec}
\newcommand\filmgamma{\gammavec}
\newcommand\filmbeta{\betavec}
\newcommand{\score}{\rho}

\DeclareMathOperator{\diag}{diag}

\newcommand\dimIn{n_{\mathrm{z}}}

\newtheorem{remark}{Remark}

\definecolor{tabblue}{RGB}{31,119,180}
\definecolor{taborange}{RGB}{255,127,14}
\definecolor{tabgreen}{RGB}{44,160,44}
\definecolor{tabred}{RGB}{214,39,40}
\definecolor{tabpurple}{RGB}{148,103,189}
\definecolor{tabbrown}{RGB}{140,86,75}
\definecolor{tabpink}{RGB}{227,119,194}
\definecolor{tabgray}{RGB}{127,127,127}
\definecolor{tabolive}{RGB}{188,189,34}
\definecolor{tabcyan}{RGB}{23,190,207}

\newcommand\dimHeads{d}
\newcommand\dimW{n_w}
\newcommand\epsbf{\boldsymbol{\varepsilon}}
\newcommand\backbone{\boldsymbol{\varphi}}

\newcommand{\fakepar}[1]{\vspace{0.5em}\noindent\textbf{#1}\hspace{0.5em}}

\newcommand\modelname{\texttt{VcVBLL}\xspace}

\newcommand\basemodelname{\texttt{VBLL}\xspace}

\newcounter{todo}
\makeatletter

\newcommand\listtodoname{List of todos}
\newcommand\listoftodos{\section*{\listtodoname}\@starttoc{tod}}
\makeatother

\newcommand{\ie}{i\/.\/e\/.,\/~}
\newcommand{\eg}{e\/.\/g\/.,\/~}
\newcommand{\cf}{cf\/.\/~}

\begin{document}

\maketitle
\thispagestyle{empty}
\pagestyle{empty}

\begin{abstract}
Agile control of robotic systems often requires anticipating how the environment affects system behavior.
For example, a driver must perceive the road ahead to anticipate available friction and plan actions accordingly.
Achieving such proactive adaptation within autonomous frameworks remains a challenge, particularly under rapidly changing conditions.
Traditional modeling approaches often struggle to capture abrupt variations in system behavior, while adaptive methods are inherently reactive and may adapt too late to ensure safety.
We propose a vision-conditioned variational Bayesian last-layer dynamics model that leverages visual context to anticipate changes in the environment. 
The model first learns nominal vehicle dynamics and is then fine-tuned with feature-wise affine transformations of latent features, enabling context-aware dynamics prediction.
The resulting model is integrated into an optimal controller for vehicle racing.
We validate our method on a Lexus LC500 racing through water puddles.
With vision-conditioning, the system completed all 12 attempted laps under varying conditions.
In contrast, all baselines without visual context consistently lost control, demonstrating the importance of proactive dynamics adaptation in high-performance applications.%
\end{abstract}
\vspace{2pt}

\noindent Video: \url{https://youtu.be/m-gcFK0moI8}.

\section{Introduction}

Accurate dynamics models are key to agile robotic autonomy, such as in scenarios like emergency avoidance or racing, where vehicles operate at the limits of handling~\cite{dallas2023hierarchical}. 
At these extremes, even minor modeling errors can cascade into catastrophic failure~\cite{sasha2025first,lew2025risk}, making precise prediction of vehicle behavior essential for both safety and performance.

Traditional vehicle dynamics models rely on simplified assumptions about road conditions and tire behavior~\cite{svendenius2007tire,carlson2003nonlinear,berntorp2014particle}. 
These assumptions hold in nominal settings but often fail when faced with significant environmental variations, especially when driving at the limit.
For example, transitioning between dry and wet pavement significantly alters friction, leading to substantially different vehicle dynamics~\cite{djeumou2024one}.
Learning-based approaches seek to compensate for such effects using online adaptation of a learned dynamics model~\cite{ding2024drifting,sasha2025first,suminaka2025adaptable,kabzan2019learning,hewing2019cautious}. However, adaptation is reactive and often too slow to prevent instability near handling limits~\cite{sasha2025first}, 
motivating a shift toward anticipatory models that utilize exteroceptive sensing to adjust control strategies proactively.

\begin{figure}[t!]
    \centering
    \hspace*{-3pt}%
    \begin{tikzpicture}
        \node[anchor=south west] at (0,0) {\includegraphics[clip, trim={1cm, 3.3cm, 6.9cm, 3.5cm}, width=\columnwidth]{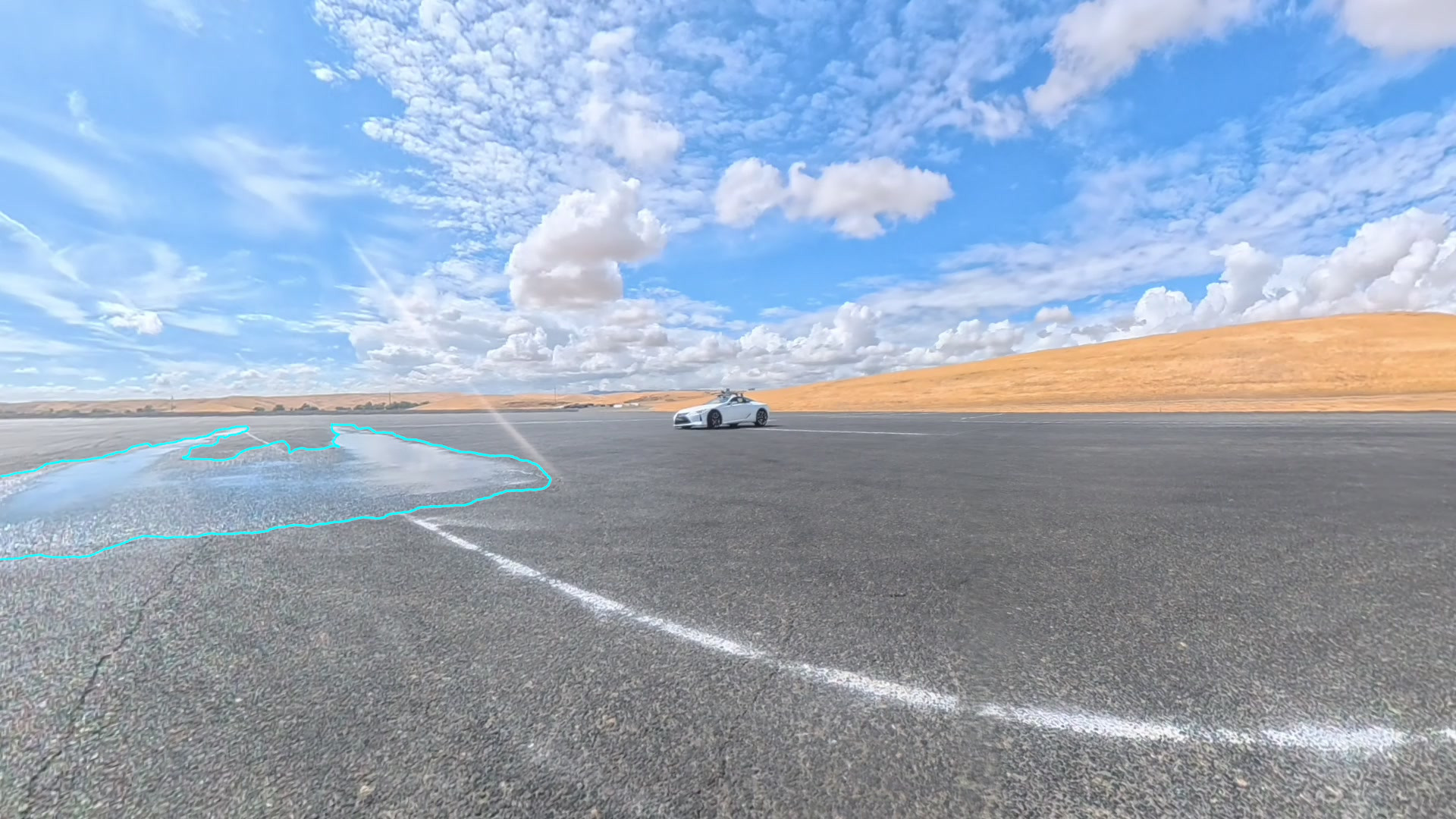}};
        \draw[line width=1pt, fill=taborange, fill opacity=0.1, taborange] (7.5,1.65) -- (2,2.55) -- (0.15,2.55) -- (0.15,0.15) -- (4.,0.15) -- (7.5,1.65);
        \node[anchor=north west] at (0,0.15) {\includegraphics[clip, trim={0.4cm, 2cm, 10cm, 7cm}, width=\columnwidth]{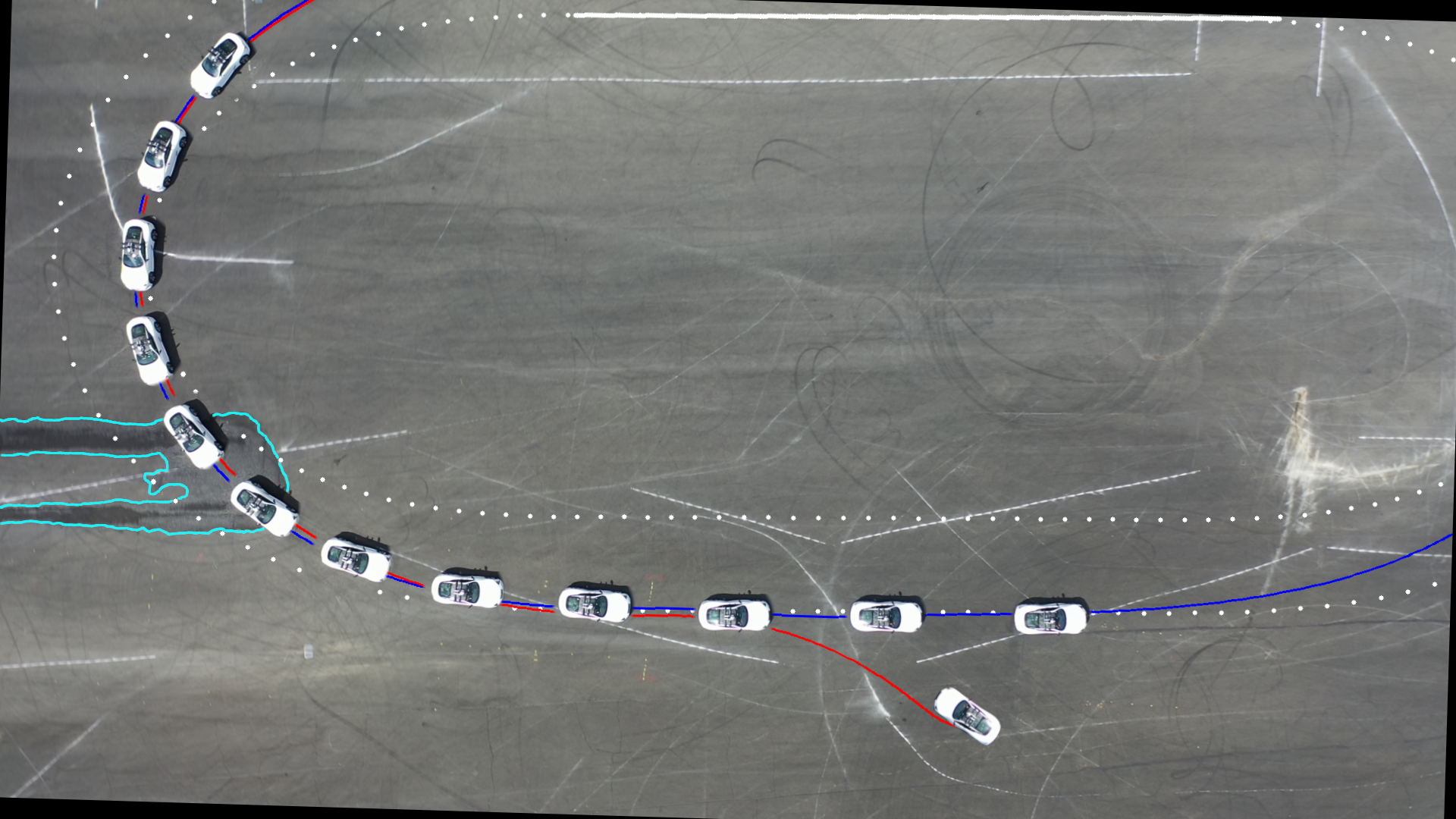}};
        \draw[thick, fill=taborange, fill opacity=0.1, taborange] (0.95,-0.8) -- (0.13,-2.2) -- (0.13,-4.35) -- (3.5,-4.35)-- (1.05,-0.8);
        
        \draw[fill=white, fill opacity=0.4, draw=black!20, line width=1pt, rounded corners=3pt] 
            (5.53,-2) rectangle (8.65,-0.1);
        
        \node[font=\small] at (7.5,-1.08) {\begin{tabular}{l}{\color{black}Without vision}\\ {\color{black}With vision}\\{\color{black}Camera FOV}\\{\color{black}Lateral limits }\\{\color{black}Water } \end{tabular}};
        \draw[line width=1.5pt, cyan!50!white] (5.8,{-1.75+0*0.353}) -- ++(0.5,0);
        \draw[line width=1.5pt, white, dotted] (5.8,{-1.75+1*0.353}) -- ++(0.5,0);
        \draw[line width=1.5pt, taborange] (5.8,{-1.75+2*0.353}) -- ++(0.5,0);
        \draw[line width=1.5pt, blue] (5.8,{-1.75+3*0.353}) -- ++(0.5,0);
        \draw[line width=1.5pt, tabred] (5.8,{-1.75+4*0.353}) -- ++(0.5,0);
    \end{tikzpicture}
    \vspace{-25pt}
    \caption{A learned vehicle model conditioned on visual context is used for predictive control. With the visual context (orange), the car autonomously races through the water (blue), while models void of context spin out (red).}
    \vspace{-1em}
    \label{fig:hero}
\end{figure}
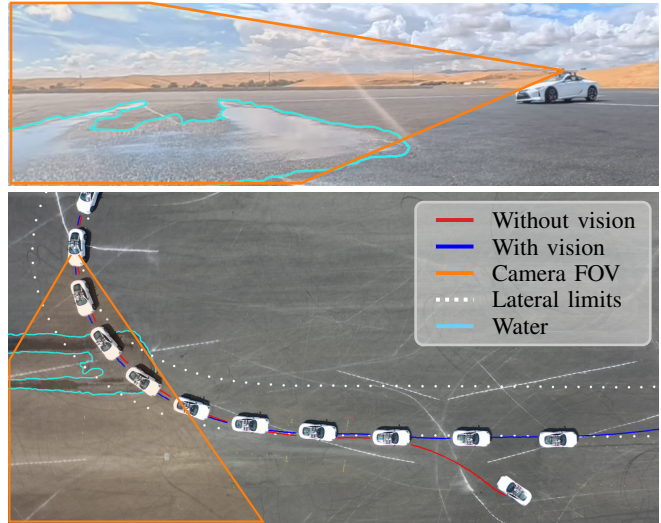

While recent learned models achieve strong performance using proprioceptive data~\cite{djeumou2023autonomous,ding2024drifting,djeumou2024one,sasha2025first}, they are often blind to the environmental context.
Incorporating exteroceptive cues from vision promises a shift from reactive to anticipatory dynamics modeling.
To this end, we introduce \modelname, a vision-conditioned variational Bayesian last-layer~(VBLL) dynamics model.
This novel model integrates visual context, extracted from forward-facing cameras, directly into the VBLL dynamics representation.
The learned conditioning on visual information allows a controller to proactively adjust its actions before encountering challenging conditions.
While this letter emphasizes racing as a challenging problem, the proposed concepts readily generalize to other applications.

\fakepar{Contributions.}
We propose a vision-conditioned VBLL dynamics (\modelname) model to capture changing vehicle dynamics arising from varying road surface properties (\eg wet vs. dry conditions) through visual information. 
Specifically:
\begin{itemize}
    \item
    With \modelname we present the first VBLL-based dynamics model conditioned on visual information.
    To address the scarcity of data with environmental changes (\eg water–tire interactions), we introduce a two-stage fine-tuning procedure. This approach modulates the dynamics representation of a base VBLL trained on nominal driving data with a vision-conditioning path learned from limited wet-surface data. The resulting model yields interpretable adjustments in the predictive distribution.
    \item
    We integrate the \modelname dynamics model within a model predictive control (MPC) framework to achieve proactive, vision-informed autonomous racing on a Lexus LC500 in hardware experiments.
    Our model enables high-speed autonomous racing through changing conditions where all proprioceptive-only baselines spin out.
\end{itemize}

\fakepar{Organization.} We review related work in Sec.~\ref{sec:related} and present preliminaries on vehicle modeling and optimal control in Sec.~\ref{sec:preliminaries}. The proposed \modelname model is developed in Sec.~\ref{sec:method} and the main results are presented in Sec.~\ref{sec:results}.

\section{Related Work}\label{sec:related}

\fakepar{Autonomous Driving at the Limit.}
High-performance driving research has focused on vehicle modeling and optimal control for racing~\cite{lew2025risk,williams2016aggressive,spielberg2021neural,subosits2021impacts,dallas2021terrain,dallas2023hierarchical} and drifting~\cite{weber2023modeling,djeumou2023autonomous,ding2024drifting,djeumou2024one}.
In these works, MPC~\cite{garcia1989model} is used to track a reference trajectory using a first-principles vehicle model~\cite{riekert1940fahrmechanik,fiala1954seitenkraften,svendenius2007tire,weber2023modeling}, often refined via system identification~\cite{ljung1998system} with a fixed parametric form~\cite{dikici2025learning}, parameter estimation~\cite{lew2025risk}, or online adaptation~\cite{dallas2023hierarchical}.
To capture more complex dynamics, neural network (NN)~\cite{spielberg2021neural} and Gaussian process~\cite{kabzan2019learning,jain2021bayesrace} models have been explored.
Recent work also demonstrates that diffusion-generated parameters can enable complex drifting~\cite{djeumou2024one}, albeit at a higher computational cost.
Other approaches include Koopman-based models that lift dynamics into a higher-dimensional space of observables~\cite{proctor2018generalizing,suminaka2025adaptable}. 
Crucially,  
these approaches do not consider proactive adaptation of the dynamics model based on exteroceptive information.
Still, parametric NNs offer a distinct advantage as a model class: their flexibility enables conditioning on visual cues via mechanisms like FiLM~\cite{perez2018film}.
Such conditioning has proven effective in vision-based manipulation~\cite{chi2025diffusion}, motivating learning exteroceptive-informed NN features from data.

\fakepar{Bayesian Last-Layer Models.}
Bayesian last-layer (BLL) models extend parametric NNs with Gaussian last-layer weights~\cite{jain2021bayesrace,snoek2015scalable,harrison2018meta,watson2021latent,fiedler2023improved} as a light-weight Bayesian NN.
They provide Gaussian predictive posteriors and have shown promise in drifting~\cite{davydov2025first} and off-road driving~\cite{levy2025meta}.
Variational BLL (VBLL) models~\cite{harrison2024variationalbayesianlayers} jointly learn last-layer weights and noise via a variational objective, achieving promising results across various applications~\cite{harrison2024variationalbayesianlayers,brunzema2025bayesian}.
Therefore, we adopt VBLLs as our base architecture, applying this state-of-the-art model class to dynamics learning for the first time.

\fakepar{Vision-Conditioned Control.}
Much work has been done on end-to-end driving, directly learning a mapping from pixels to control signals~\cite{bojarski2016end}. 
Numerous reinforcement learning approaches have been proposed for racing~\cite{cai2021vision,balaji2019deepracer}, focusing on learning an NN policy.
Also in MPC, vision-conditioning has been applied to infer cost \cite{drews2017aggressive,xiao2022learning} or derive constraints for obstacle avoidance \cite{drews2017aggressive}, but not to condition a dynamics model.
Although recent methods learn MPC dynamics within an RL framework for aggressive driving, they do not incorporate vision~\cite{adabag2025differentiable}.
In mobile robotics, visual foundation models have been used to estimate disturbances for a nominal physics-based dynamics model~\cite{lupu2024magic}, and perceptive dynamics models have been proposed for legged robots using exteroceptive inputs to enable safe planning~\cite{roth2025learned} with MPPI.
To our knowledge, our approach is the first to 
explicitly condition a dynamics model on vision with downstream use in an optimal controller for proactive racing.
Our results show that this capability can be the difference-maker when racing autonomously in varying friction conditions.

\section{Preliminaries}\label{sec:preliminaries}

\begin{figure}[t!]
    \centering
    \input{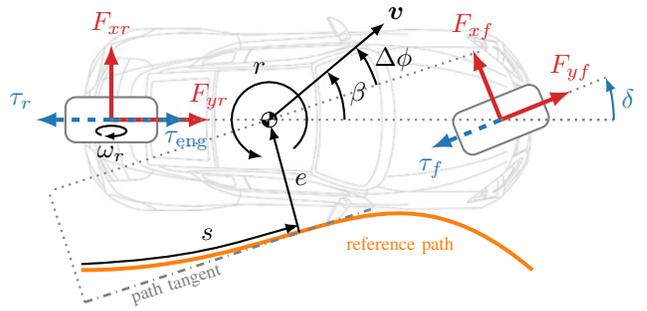}
    \vspace{-20pt}
    
    \caption{Vehicle geometry with the bicycle model states $\xvec$ (black), control inputs $\uvec$ (blue), tire forces (red), and the reference path $\xvec_{\mathrm{ref}}(s)$ (orange).}
    \vspace{-1em}
    \label{fig:geometry}
\end{figure}

Vectors are denoted by $\xvec\in \mathbb{R}^{n}$, with $x_i$ being the $i${th} element of $\xvec$. Matrices are indicated in bold, $\XX$, and the element on row $i$ and column $j$ of $\XX$ is $[\XX]_{ij}$. We let
\begin{equation*}
    \ubar{\xvec} \sim \Gaussian(\xvec|\mvec,\Pbf)\propto |\Pbf|^{-1/2}\exp(-\tfrac{1}{2}(\xvec-\mvec)^{\top}\Pbf^{-1}(\xvec-\mvec))
\end{equation*}
indicate that the random variable $\ubar{\xvec}$ is Gaussian distributed with mean $\mvec$ and covariance $\Pbf$. Similarly,  we let
\begin{equation*}
    \ubar{x} \sim \mathrm{IG}(x|\alpha,\beta)\propto x^{-\alpha-1}\exp(-\beta x^{-1}), 
\end{equation*}
indicate that $\ubar{x}$ is inverse Gamma distributed.  We write the expectation of $\ubar\xvec\sim p_{\ubar\xvec}(\xvec)$ as $\mathbb{E}_{ \ubar\xvec \sim p_{\ubar\xvec}(\xvec)}[\ubar\xvec]=\int \xvec p_{\ubar\xvec}(\xvec)\der\xvec$ compactly as $\mathbb{E}_p[\xvec]$. In this notation, the KL-divergence between the densities $p$ and $q$ is $\mathrm{KL}(p\|q) = \mathbb{E}_p[\log(p(\xvec)/q(\xvec))]$.

\fakepar{Physics-based models.} In~\cite{spielberg2021neural,davydov2025first,lew2025risk}, it is shown that a well-calibrated single-track model~\cite{riekert1940fahrmechanik} with Fiala tire-forces~\cite{fiala1954seitenkraften,svendenius2007tire} enables racing at the limits of handling (see Fig.~\ref{fig:geometry}). 
Thus, we use similar inputs for our learned model, and define the state $\states=(\states_{1}, \states_{2})$ and control input $\controls$ as
\begin{subequations}
    \begin{align}
        \states_{1} &= (r, v, \beta, \omega_r)^{\top}\in\Real^4,\\
        \states_{2} &= (e, \Delta\phi, s)^{\top}\in\Real^3,\\
        \controls &= (\delta, \tau_{\text{combined}}, \tau_{f})^{\top}\in\Real^3,
    \end{align}
\end{subequations}
where $r$ (rad/s) is a yaw rate, $v$ (m/s) is the longitudinal velocity, $\beta$ (rad) is a side-slip angle, and $\omega_r$ (rad/s) is an average rear wheel speed, $e$ (m) is a lateral path distance, $\Delta\phi$ (rad) is a deviation with respect to a racing line, and $s$ (m) is a path distance (see Fig.~\ref{fig:geometry}). Furthermore, $\delta$ (rad) is the steering angle, $\tau_{\text{combined}}$ (Nm) is the combined rear torque, and $\tau_{f}$~(Nm) are the front brake torques. The dynamics of the vehicle can then be written as a cascaded system
\begin{subequations}
\begin{align}
\dot{\states}_1(t) &= \fvec_1(\states_1(t), \uvec(t)),\label{eq:learned_dynamics}\\
\dot{\states}_2(t) &= \fvec_2(\states_1(t), \states_2(t), \uvec(t)).
\end{align}
\end{subequations}
In this representation, $\fvec_1$ encodes assumptions about the tire-force model and is defined by parameters that are estimated or manually tuned (see the appendix), whereas $\fvec_2$ is known from the car's kinematics. Indeed, from Fig.~\ref{fig:geometry}, we obtain
\begin{subequations}
\begin{align}
\dot{e}(t) &= v(t)\sin(\Delta\phi(t)), \\
\Delta\dot{\phi} &= \dot\beta(t) + r(t) - \frac{\kappa_{\textrm{ref}}(t)v(t)\cos(\Delta\phi(t))}{(1-\kappa_{\textrm{ref}}(t)e(t))}, \\
\dot{s}(t) &= \frac{v(t)\cos(\Delta\phi)}{(1-\kappa_{\textrm{ref}}(t)e(t))},
\end{align}
\end{subequations}
where $\kappa_{\mathrm{ref}}$ is the signed curvature of the reference path.  The term $\fvec_1$ contains all the epistemic uncertainty of the vehicle model, and is the focus of our learning based methods. To embed this model in a predictive control framework, we reparametrize the model in the path variable $s$, as 
\begin{subequations}
\begin{align}\label{eq:reparametrize}
    \frac{\mathrm{d}}{\mathrm{d}s}\xvec(s) &= \frac{1}{\dot{s}}\begin{bmatrix}
        \fvec_1(\states_1(s), \uvec(s))\\
        \dot{e}(s)\\
        \dot{\Delta\phi}(s)\\
        1
    \end{bmatrix}\triangleq \boldsymbol{f}(\states(s), \controls(s)).
\end{align}
\end{subequations}
We then discretize the dynamics with a constant path step $\Delta s = \SI{3}{\meter}$ and integrate the ODE with explicit Euler as
\begin{equation}\label{eq:discdynins}
    \states_{k+1} = \states_k + \Delta s  \boldsymbol{f}(\states_{k}, \controls_{k}).
\end{equation}

\fakepar{Optimal Control.} In the context of racing, the objective is to minimize lap time. Following~\cite{lew2025risk}, we penalize state-control deviations from a pre-computed reference trajectory $(\states_{\text{ref}}(s), \controls_{\text{ref}}, \kappa_{\text{ref}}(s))_{s>0}$ and rapid changes in control inputs to improve robustness and numerical stability. We define the linear time cost and quadratic stage costs as:
\begin{subequations}
\begin{align}
\ell_T(\states) \!&=\! \lambda_T \, t, \\ 
\ell(\states,\!  \controls,\!  \controls_{-}) \!&=\! \|\states \!- \!\states_{\text{ref}}\|_{\Q}^2\! +\! \|\controls \!- \!\controls_{\text{ref}}\|_{\R}^2 \!+ \!\frac{\|\controls \!- \! \controls_{-}\|_{\T}^2}{(\Delta s)^2},
\end{align}
\end{subequations}
where $\| \vvec \|_{\Pbf}^2 = \vvec^\top \Pbf \vvec$, and $(\Q, \R, \T)$ are diagonal weight matrices with small entries compared to terminal time cost $\ell_T$ (see the appendix). The term $\|\controls -\controls_{-}\|_{\T}^2 / (\Delta s)^2$ penalizes control rates, where $\controls_{-}$ is the previous control input. With these costs, we formulate the optimal control problem (OCP):
\begin{align}\label{eq:OCP}
\min_{\substack{\states_{0:N}\\\controls_{0:N-1}}} \quad & \ell_T(\states_N) + \textstyle\sum_{k=0}^{N-1} \ell(\states_k, \controls_k, \controls_{k-1}) \\
\text{s.t.} \quad &  \F(\states_k, \controls_k, \states_{k+1} \mid \context_t) = \mathbf{0}, \hspace{35pt}  k \in [0, N-1] \notag\\
& (\states_0, \controls_0) = (\states_{\text{init}}, \controls_{\text{init}}), \notag\\
& \controls_{\min} \leq \controls_k \leq \controls_{\max}, \hspace{61pt} k \in [0, N-1] \notag\\
& e_{\min,k} \leq e_k \leq e_{\max,k},  \hspace{54pt} k \in [0, N]\notag
\end{align}
where $\F$ represents the combined dynamics given the current context $\context_t$, enforcing~\eqref{eq:discdynins}.
This formulation results in a context-dependent optimal control problem that is solved in a receding horizon fashion. The car is actuated with the optimal controls $\controls^{\star}_{0:N-1}$ linearly interpolated to the current position of the car in $s$. Next, we describe how to express $\F$ using VBLLs.

\def\sep{0.1}
\def\boxwidth{3.45}
\def\boxheight{3}
\begin{figure*}[t!]
    \centering
    \includegraphics[width=\textwidth]{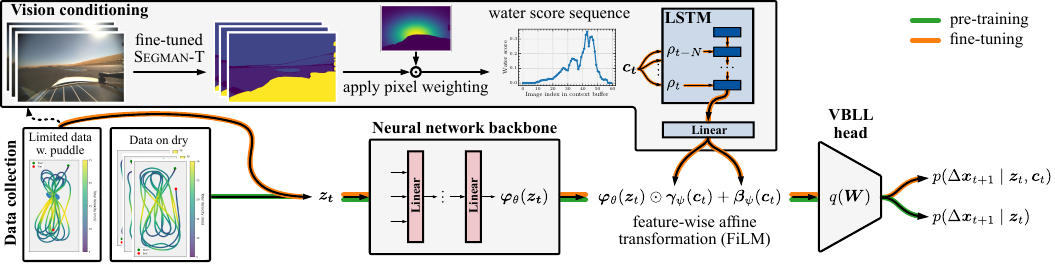}
    \vspace{-20pt}
    
    \caption{
    \textit{Conceptual overview of \modelname and its training.} 
    We begin by training a base VBLL dynamics model (green path).
    To incorporate vision-based conditioning, we first extract semantic classes using a fine-tuned \textsc{Segman-t} backbone from the limited available data with water interaction, from which we compute a short \SI{2}{\sec} time series of past water scores corresponding to $\context_t$. 
    In a subsequent fine-tuning phase, this time series is encoded with an LSTM and used in a FiLM conditioning path to obtain our vision-conditioned \modelname dynamics model (orange path).}
    \vspace{-1em}
    \label{fig:overview}
\end{figure*}

\section{VBLLs with FiLM Conditioning}\label{sec:method}

We begin in Sec.~\ref{sec:VBLLbase} by describing how to learn a base VBLL dynamics model from state transitions only. 
Sec.~\ref{sec:backbone} explains how to extract visual context from limited data, and Sec.~\ref{sec:film} details how to condition the VBLL model on this context in a fine-tuning step with limited data (see Fig.~\ref{fig:overview}).

\subsection{Learning a VBLL vehicle model}\label{sec:VBLLbase}

Following common practice in dynamics learning \cite{deisenroth2011pilco,ding2024drifting}, we train a VBLL model to predict state differences rather than absolute states, which has been shown to improve numerical stability~\cite{ding2024drifting}. 
We learn the dynamics~\eqref{eq:learned_dynamics} in time coordinates and then convert to path coordinates using~\eqref{eq:reparametrize}. Specifically, for a time step $\Delta t>0$, we model
\begin{equation}\label{eq:statediff}
    \Delta \states_{t} \coloneqq \states_{1, t+\Delta t} - \states_{1, t}  = \fvec_{\theta} (\inputs_t),
\end{equation}
where $\fvec_{\theta}$ denotes the learned dynamics function parameterized by $\theta$, and $\inputs_t = [\states_{1, t}, \controls_t, \controls_{t+\Delta t}]\in\Real^{\dimIn}$ contains the relevant states and controls at time $t$ along with the controls at the next time step.\footnote{The subscript $t$ is a time (continuous) and $k$ is a path coordinate (discrete).}
Selecting a $\Delta t$ to approximately match the discrete path step $\Delta s$ at the expected average velocity will later allow us to directly integrate the learned model using the discretization scheme in \eqref{eq:discdynins}.

\begin{remark}
As in~\cite{lew2025risk,sasha2025first}, we include the control input at the next time step in the dynamics model inputs, consistent with our MPC implementation that interpolates the control sequence in $s$ at runtime.
\end{remark}

Given the training target description in \eqref{eq:statediff}, we now define the VBLL model architecture that learns to predict these state differences in a supervised manner.
We define a VBLL model with $\dimHeads$ independent heads as 
\begin{equation}\label{eq:model}
   \fvec_\theta (\zvec_k) = \backbone_{\theta} (\zvec_k)^\top \W = \backbone_{\theta} (\zvec_k)^\top [\wvec_1, \dots , \wvec_{\dimHeads}],
\end{equation}
where $\backbone_{\theta}:\Real^{\dimIn}\mapsto \Real^{\dimW}$ are learned features and $\wvec_i \in \Real^{\dimW}$ are Gaussian random variables associated with each VBLL prediction head. Sampling the state difference~\eqref{eq:statediff} in practice is noisy, and we therefore define a measurement model
\begin{equation}\label{eq:meas}
\yvec_k = \fvec_\theta(\zvec_k) + \epsbf_k,
\end{equation}
where $\epsbf_k$ is independent Gaussian zero-mean noise with covariance $\boldsymbol{\Sigma}=\mathrm{diag}(\Sigma_1, \cdots, \Sigma_{\dimHeads})$. To infer the backbone parameters, weight distribution, and noise statistics, we follow~\cite{harrison2024vbll} and assume a zero-mean Gaussian prior with covariance $p(\wvec_i)=\mathrm{N}(\wvec_i; \Z, \xi^{\circ}_i\I)$ for the $i$th head.
The measurements in~\eqref{eq:meas} are Gaussian given~\eqref{eq:model}.
As the Gaussian is its own conjugate prior, we
define the posterior
\begin{equation}\label{eq:posterior}
    q(\W) = 
    \prod_{i=1}^{\dimHeads}q(\wvec_i)=\prod_{i=1}^{\dimHeads}\Gaussian(\wvec_i ; \mvec_i, \Sbf_i),
\end{equation}
and let $\Sbf_i = \Lbf_{i}\Lbf_{i}^\top$ where $\Lbf_{i} = \N_{{i}} + \diag(\exp(\myvec{d}_{i}))$ is lower-triangular and $\mathrm{diag}(\N_{{i}})=\Z$.  This ensures that $\Sbf_i$ is positive definite without imposing explicit constraints during training. For ease of notation, we define the learnable parameters of the VBLL heads as $\eta = \{\mvec_i, {\N}_{i}, \myvec{d}_i, \Sigma_i \}_{i=1}^{\dimHeads}$.

\begin{remark}
This approach differs from the BLL in~\cite{sasha2025first} in that $\eta$ includes the unknown noise statistics $\{\Sigma_i\}_{i=1}^d$, giving the model more flexibility. 
As in \cite{sasha2025first}, our model allows for online adaptation, which is one of our baselines in Sec.~\ref{sec:results}.
\end{remark}

\fakepar{Losses.} Let $\mathcal{D} = \{(\inputs_j, \yvec_j)\}_{j=1}^{|\mathcal{D}|}$ denote our training dataset, where each sample consists of inputs $\inputs_j$ and the corresponding measured state differences $\yvec_{j}$.
We standardize all inputs and outputs to have a zero mean and unit variance, thereby improving training stability and convergence.
We seek to maximize the evidence lower bound (ELBO) derived in \protect{\cite[Theorem 1]{harrison2024vbll}}. Given the weights in~\eqref{eq:posterior} the noise in~\eqref{eq:meas} are both independent across the VBLL heads, we get

\begin{align}
    \mathcal{L}_{\text{ELL}}(\eta, \theta) =& \frac{1}{|\mathcal{D}|} \sum_{j=1}^{|\mathcal{D}|}\sum_{i=1}^{d} \Big[ \log \Gaussian(y_{ji} | \backbone_{\theta}(\inputs_j)^\top \mvec_i, \Sigma_i) \nonumber \\
    & - \frac{1}{2} \backbone_{\theta}(\inputs_j)^\top \Sbf_i \backbone_{\theta}(\inputs_j)  \Sigma_i^{-1}\Big],
\end{align}
as a lower bound on the expected log-likelihood (ELL) of $\Dcal$.
Following~\cite{harrison2024vbll}, we also introduce a regularizer
\begin{align}
    \mathcal{R}(\eta, \theta) =& \frac{1}{|\mathcal{D}|}\sum_{i=1}^{\dimHeads} \Big[ - \text{KL}(q(\wvec_i)||p(\wvec_i)) \nonumber \\
     &-\log\Gaussian(\theta | \Z, \gamma^{\circ}\I)  - \log \mathrm{IG}(\Sigma_i | \alpha_i^{\circ}, \beta_i^{\circ}) \Big].
\end{align}
weighted by $\lambda_{\text{reg}}>0$ to stabilize training and regularize the densities toward suitable priors defined by $\{\alpha_i^{\circ},\beta_i^{\circ},\gamma^{\circ}, \xi_i^{\circ}\}_{i=1}^{\dimHeads}$.
Combining both parts, we formulate and solve the following optimization problem
\begin{equation}\label{eq:base_vbll_loss}
    \bar\theta, \bar\eta = \arg\min_{\theta, \eta} \underbrace{\left[-\mathcal{L}_{\text{ELL}}(\eta, \theta) + \lambda_{\text{reg}} \mathcal{R}(\eta, \theta)\right]}_{\eqqcolon \mathcal{L}(\eta, \theta)},
\end{equation}
with additional details on the priors, specifications of the MLP backbone $\backbone$, and considered regularizers given in the Appendix. 
With the learned parameters, we obtain a posterior predictive given an input at time $t$ as
\begin{equation}
    y_{ti} \mid \inputs_t  \sim \Gaussian (\backbone_{\bar\theta}(\inputs_t)^\top \bar\mvec_i,  \backbone_{\bar\theta}(\inputs_t)^\top \bar\Sbf_i \backbone_{\bar\theta}(\inputs_t) + \bar\Sigma_i).
\end{equation}

To use this model in the intended racing application, we use the maximum a posteriori (MAP) estimate of the learned state difference in the ODE~\eqref{eq:learned_dynamics} of the states $\xvec_1$, and let
\begin{equation}\label{eq:meanpred}
\dot{\xvec}_1(t) \approx \backbone_{\bar\theta}(\inputs_t)^{\top}[\bar\mvec_1, \cdots, \bar\mvec_{\dimHeads}] / \Delta t.
\end{equation}
This approach allows us to use the curvilinear coordinates and reparametrize the dynamics in path distance~\eqref{eq:reparametrize}, discretize the dynamics in $s$~\eqref{eq:discdynins}, and formulate the OCP~\eqref{eq:OCP} in Sec.~\ref{sec:preliminaries}.

\subsection{Extracting Contextual Visual Information}\label{sec:backbone}

\fakepar{Backbones.} To extract meaningful visual context for the learned model in Sec.~\ref{sec:VBLLbase}, we use \segman~\cite{fu2025segman} to segment images captured on the car. We also evaluated \segnext~\cite{guo2022segnext} and \textsc{SegFormer}~\cite{xie2021segformer}, but found \segman superior in both speed and accuracy for our driving application.
We use the tiny model and train it from scratch on the public \goose dataset~\cite{mortimer2024goose}, which contains 10k images of rural and off-road driving in Germany. This setting better matches racing conditions, which are closer to rural than urban environments, yielding better embeddings.
To further improve performance, we fine-tune the model on 500 manually labeled images from the California test location used in our experiments.
We retain the 64 closed-set classes of the \goose dataset, represented by unique integers $\Ccal = \{1, \ldots, 64\}$. At inference, this yields a segmented image $\hat{I}_k \in [|\mathcal{C}|]^{W\times H}$ of size $(W, H) = (1440, 928)$. Example outputs are shown in Fig.~\ref{fig:overview}.

\fakepar{Weighting.} To extract semantically meaningful surface information from this output, we compute a spatially weighted water presence score. This score emphasizes the region of the image most relevant to the vehicle's trajectory. We define a Gaussian weighting mask over the image domain. At pixel coordinates $(w,h) \in [W] \times [H]$, the weight is given by
\begin{equation}
w(h, w) = \exp\left( -\left[ \frac{(h - \mu_h)^2}{2\sigma_h^2} + \frac{(w - \mu_w)^2}{2\sigma_w^2} \right] \right),
\end{equation}
where $\mu_h \in [H]$ and $\mu_w \in [W]$ and $\sigma_h, \sigma_w > 0$ control its spread in the vertical and horizontal directions, respectively. This approach focuses attention on a region of interest, such as the road area directly ahead of the vehicle, and down-weights potential outliers on the edge of the image (see Fig.~\ref{fig:overview}).

\fakepar{Context.} Without loss of generality, we consider two classes:
\begin{itemize}
   \item $c_{\texttt{w}} \in \mathcal{C}$: the class label corresponding to water;
   \item $c_{\texttt{a}} \in \mathcal{C}$: the class label corresponding to asphalt.
\end{itemize}
Furthermore, let $\Omega \subset [W] \times [H]$ denote the set of pixels that are not occluded (i.e., not part of the vehicle body), for which the weight is non-zero. We then define a ``water score'' as
\begin{equation}\label{eq:scored}
    \score_t = 
    \frac{
       \sum_{(w, h) \in \Omega} w(h, w) \cdot \mathbf{1}\{\hat{I}_k = c_{\texttt{w}}\}
    }{
       \sum_{(w, h) \in \Omega} w(h, w)\! \cdot\! \left[
           \mathbf{1}\{\hat{I}_k \!= c_{\texttt{w}}\}\! +\! \mathbf{1}\{\hat{I}_k \!= c_{\texttt{a}}\}
       \right]
    },
\end{equation}
where $\mathbf{1}\{\cdot\}$ denotes the indicator function.
This ratio provides a normalized measure of the prevalence of water relative to asphalt in the spatial region emphasized by the Gaussian. In general, if there are more classes of relevance, such as ice or gravel, we can easily extend~\eqref{eq:scored} to a vector-valued score, with the relative prevalence of each class of interest. In the following, we provide the context $\context_t = (\score_{t-N}, \dots, \score_t)$ of the current surface condition to the learned dynamics models.

\subsection{Vision-based Conditioning using FiLM}
\label{sec:film}

To incorporate contextual visual information into our learned dynamics model, we employ Feature-wise Linear Modulation (FiLM) \cite{perez2018film}. This approach allows the visual context to adaptively modulate the backbone network's features through learned affine transformations. FiLM relies on a conditioning function $g_{\psi}(\cdot)$ that maps the contextual input to modulation parameters. We implement this function as a Long Short-Term Memory (LSTM) network~\cite{hochreiter1997long}, with
\begin{equation}
    (\filmgamma_{\psi}(\context_t), \filmbeta_{\psi}(\context_t)) = g_{\psi}(\context_{t}),
\end{equation}
where $g_{\psi}$ is parameterized by LSTM weights $\psi$, and the factors $\filmgamma(\context_t)$ and shift parameters $\filmbeta(\context_t)$ have the same dimensionality as the backbone features (see Sec.~\ref{sec:backbone}). While it is possible to use simple MLPs, the LSTM is chosen to better capture temporal dependencies in the context $\context_t$.

\fakepar{Modulation.} The backbone features of the VBLL are modulated as in~\cite{perez2018film}, with an element-wise affine transformation
\begin{equation}
   \tilde{\features}_t \coloneqq \features^{\text{FiLM}}_{\{\theta, \psi\}} (\inputs_t, \context_{t}) = \features_\theta (\inputs_t) \odot \filmgamma_\psi(\context_{t}) + \filmbeta_\psi(\context_{t}),
\end{equation}
where $\odot$ denotes element-wise multiplication. This formulation enables the visual context to selectively amplify or suppress different feature dimensions based on the observed surface conditions, allowing the model to adapt its internal representations to account for varying road surface properties.
Finally, we formulate the conditional posterior predictive
\begin{equation}\label{eq:filmpostpred}
    \Delta \states_{t+1, i} \mid \zvec_t, \context_{t} \sim \Gaussian\left(\tilde{\features}_t^\top \mvec_i, \tilde{\features}_t^\top \Sbf_i \tilde{\features}_t + \Sigma_i \right).
\end{equation}

\subsection{Data Collection, Training, and Fine-Tuning}

\fakepar{Data Collection.}
An expert driver operated a Lexus LC500 on a skidpad under controlled conditions to collect training data.
The nominal dataset, containing $|\mathcal{D}|=\num{12229}$ samples for training, was obtained through a series of aggressive maneuvers designed to excite the full range of the nonlinear vehicle dynamics in nominal, \ie dry, conditions (see Fig.~\ref{fig:overview}).

For the fine-tuning dataset, we introduced a water puddle on the skidpad. The driver then repeated similar aggressive maneuvers to capture transient tire-surface interactions under wet conditions (\cf Fig.~\ref{fig:overview}, bottom left). High ambient temperatures and sun exposure at the test site caused rapid evaporation, making this data collection process slow and labor-intensive, which highlights the utility of our fine-tuning approach.
The fine-tuning dataset comprises \num{6349} samples, of which only \num{1540} (approx. 24\%) contain a non-zero context vector~$\context_t$.
Since $\context_t$ spans a two-second window, the direct exposure to wet conditions is even more limited.

\fakepar{Training Procedure.}
We first train the base VBLL dynamics model on the nominal dataset by minimizing the loss in \eqref{eq:base_vbll_loss}. To enhance numerical stability, all input and output features are standardized prior to training.
The same standardization parameters are applied to the fine-tuning dataset to ensure feature consistency.
We then use this trained base VBLL model as the initialization for the subsequent fine-tuning to learn the conditional path (i.e., the orange path in Fig.~\ref{fig:overview}).

To prevent catastrophic forgetting, we augment the base loss from \eqref{eq:base_vbll_loss} with an L2-SP regularization term \cite{xuhong2018explicit}, and solve the following optimization problem using the fine-tuning dataset
\begin{equation}
    \hspace{-2pt}\bar\theta, \bar\eta, \bar\psi = \arg\min_{\theta, \eta, \psi} \left[ \mathcal{L} ( \eta, \{\theta, \psi\}) + \lambda_{\text{L2-SP}} \left\| \theta \!- \!\theta_{\text{init}} \right\|_2^2 \right],\hspace{-2pt}
\end{equation}
where $\theta_{\text{init}}$ are the parameters of the VBLL model trained with a loss~\eqref{eq:base_vbll_loss} using only dry data (green path in Fig.~\ref{fig:overview}).
We refer to the resulting finetuned model as \modelname.

\begin{figure*}[t!]
    \centering
    \includegraphics[width=\textwidth]{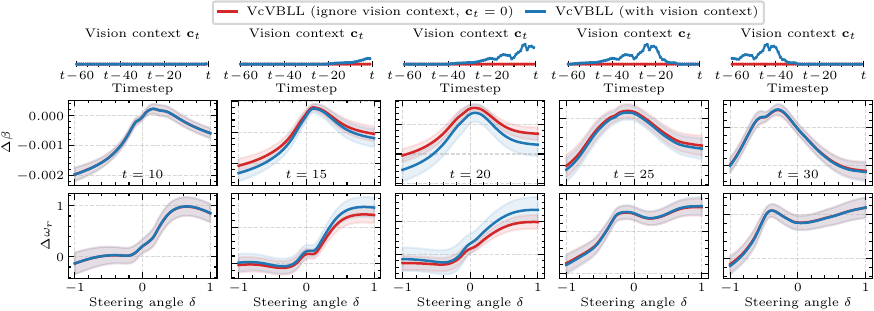}%
    \vspace{-10pt}
    
    \caption{
    \textit{Slices through the \modelname posterior when driving over a puddle.}  
    After fine-tuning, the \modelname model can effectively provide conditional prediction based on the provided visual context, as seen by the difference in posterior predictive distributions for the selected states.
    Moreover, the model’s behavior is physically plausible: a wet surface results in higher wheel speeds and increased side-slip dynamics, as well as increased uncertainty.}
    \vspace{-1em}
    \label{fig:qualitative:posterior}
\end{figure*}

\section{Results}\label{sec:results}

We test the proposed \modelname vehicle model in closed-loop MPC on a Lexus LC500.
We discuss the effect of the visual context in Sec.~\ref{sec:qualitative} and then present hardware results with baseline models, with and without the vision context, when racing through puddles in Sec.~\ref{sec:hardware}. A video of the experiments is available at: \url{https://youtu.be/m-gcFK0moI8}.

\subsection{Intuition on the Effect of Vision Conditioning}\label{sec:qualitative}

To illustrate the effect of the vision context on model predictions, we study the fine-tuned \modelname described in Sec.~\ref{sec:film}, with and without vision context.
We use segmentation outputs recorded in the car while racing through a puddle, and show the posterior predictive~\eqref{eq:filmpostpred} as a function of the steering input $\delta$ at different points in time (see Fig.~\ref{fig:qualitative:posterior}).

The \modelname predictions diverge upon detecting water (${t=15}$): with the context, the wheels are predicted to slip and spin more, and the car is also predicted to lose traction and gain more sideslip under the same actuation. Notably, the difference induced by the context is maximal after a small delay, corresponding to the time between detecting water and the vehicle entering the puddle ($t=20$). These differences in prediction persist for some time due to the tires remaining wet after having passed through the puddle ($t=25$). Over time, the impact of water on the dynamics becomes negligible ($t=30$). The effect of the vision context on the VBLL predictions thus makes sense physically. The presence of water also inflates the uncertainty
 of the predictive posterior,  which is expected as the vehicle dynamics become harder to model accurately under such transient surface conditions.

\subsection{Hardware Experiments}\label{sec:hardware}

\fakepar{Setup.} 
We validate the proposed method on the LC500 vehicle depicted in Fig.~\ref{fig:hero}. The segmentation backbone is run on an L4 GPU, and all other computations are done on a 3.30GHz Intel Xeon E-2278GE CPU. This includes evaluating the FiLM modulation and solving the OCP via sequential quadratic programming~\protect{\cite[Chapter 18]{nocedal2006numerical}} using the OSQP solver~\cite{stellato2020osqp}. The vehicle states are estimated using an OxTS system~\cite{oxts2022}, and the images are sampled from a single forward-facing Lucid Ti 3.2MP camera~\cite{triton2025} (see Fig.~\ref{fig:overview}). With this setup, the segmentation runs at approximately 25-30Hz, and is therefore separated in a different thread to not prevent the execution of the controller. Compared to the offline training data for the FiLM conditioning, which provides segmentation results at 30Hz, some frames are dropped in the experiments. Consequently, we align the vehicle states with the images and populate the missing frames by interpolating the water score. When present, the modulating FiLM signal is computed at 7Hz, while the controller is running at 62.5Hz.

\begin{figure}[t!]
    \centering
    \includegraphics[width=\columnwidth]{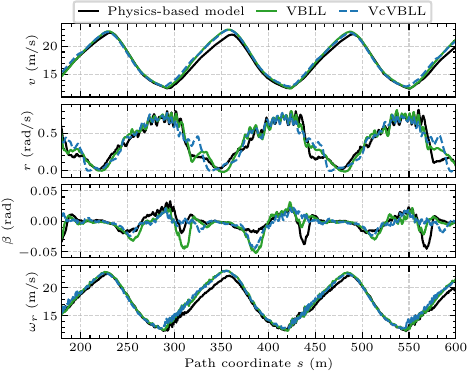}
    \vspace{-20pt}
    
    \caption{
    \textit{Performance of VBLL dynamics model on the dry track.}
    With a learned dynamics model (\basemodelname and \modelname) the MPC is able to achieve higher velocities resulting in faster lap times (\cf Table~\ref{tab:laptimes}).
    }
    \vspace{-1em}
    \label{fig:baselines}
\end{figure}

\fakepar{Racing Baselines.} To verify that the learned VBLL serves as a suitable evaluation baseline, we first test it in hardware using the deterministic minimum-time MPC formulation from~\cite{lew2025risk} (without puddles), and compare:
\begin{itemize}
    \item The \texttt{physics-based} vehicle model used in~\cite{lew2025risk};
    \item A baseline \basemodelname vehicle model, similar to that of~\cite{sasha2025first};
    \item A \modelname model derived from the baseline \basemodelname.
\end{itemize}
We consider the racing task in~\cite{lew2025risk}, and report results in Fig.~\ref{fig:baselines}. We observe that the controllers using the OCPs formulated with VBLL models better utilize the available rear tire force by applying more throttle, resulting in more rear wheel speed (see $\omega_r$), sideslip (see $\beta$), and ultimately velocity (see $v$) when exiting the corner. Performance is reduced marginally in \modelname compared to \basemodelname, but both methods still beat the \texttt{physics-based} model in lap times (see Tab.~\ref{tab:laptimes}). Next, we introduce water on the track and study the effect of vision.

\begin{figure*}[!t]
    \centering
    \includegraphics[width=\textwidth]{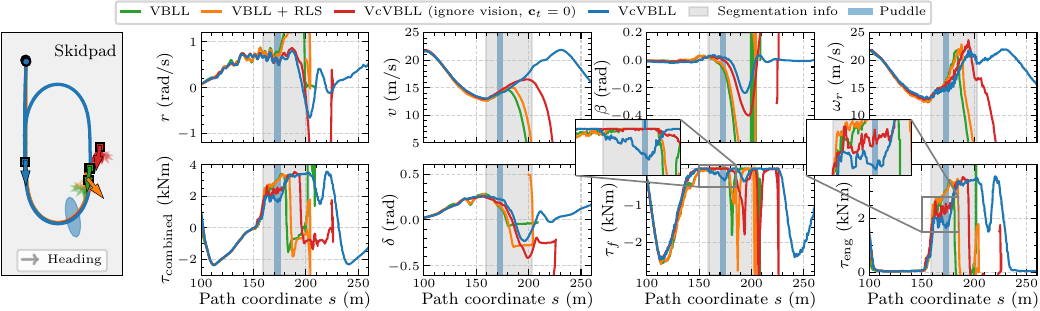}%
    \vspace{-10pt}

    \caption{Hardware experiments comparing various methods in a racing task with water placed on the exit of a turn. All methods spin out after driving through the puddle, except for the \modelname with vision context, which completes three laps before the experiment is stopped manually.}
    \vspace{-1em}
    \label{fig:hardware_res}
\end{figure*}

\begin{table}[tp!]
    \sisetup{output-decimal-marker = {.}}
    \centering
    \caption{Lap time of baseline methods for Figure~\ref{fig:baselines} without puddle.}
    \vspace*{-0.8em}

    \begin{tabular}{cccc}\toprule
     & \texttt{Physics-based} & \basemodelname  & \modelname  \\\midrule
    \textbf{Lap time} & \SI{24.53}{\sec} & \SI{24.06}{\sec} & \SI{24.24}{\sec}\\
    \bottomrule
    \end{tabular}
    \label{tab:laptimes}
    \vspace{-1em}
\end{table}

\fakepar{Racing Through Puddles.}
To study how the vision context affects closed-loop control, we place a large puddle near the exit of the first turn and replenish it with equal amounts of water between each consecutive experiment. We use the minimum-time MPC with dynamics constraints from:
\begin{itemize}
    \item The baseline \basemodelname vehicle model;
    \item The \basemodelname model with RLS adaptation, as in~\cite{sasha2025first};
    \item The \modelname model ignoring vision context;
    \item The \modelname model utilizing vision context.
\end{itemize}
The results from one of three sets of experiments are shown in Fig.~\ref{fig:hardware_res}, and indicate that the vision context is crucial for the car to drive safely through the puddle. In particular, as soon as there is a signal from the segmentation that water will be present, the controller preemptively drops the engine torque and engages the brakes, which reduces the side-slip on exit and allows the vehicle to recover and continue racing. The \modelname without vision context exhibits similar behavior in engine torque, but does not initiate the brakes and spins out as a result. These experiments were repeated three times, with the \modelname using vision context completing all 9 laps through the water successfully. We conclude that the vision context, albeit simplistic, allows for preemptive control through the MPC, and is the unambiguous difference maker between staying on track and spinning out.

\fakepar{Robustness.} To test robustness to vehicle parameter changes like road conditions and tire wear, we deployed the method twice with {(i)} different tires, {(ii)} 15$^{\circ}$C temperature difference, and {(iii)} different lighting (see Fig.~\ref{fig:lightingconditions}). 
We slightly re-tuned the parameters of the controller for all baselines, using $\lambda_T=500$ on the first occasion and $\lambda_T=200$ on the second occasion. 

\usetikzlibrary{positioning}
\begin{figure}[t!]
    \centering
    \begin{tikzpicture}
        \node[inner sep=0pt] (aug) {\includegraphics[width=0.47\columnwidth]{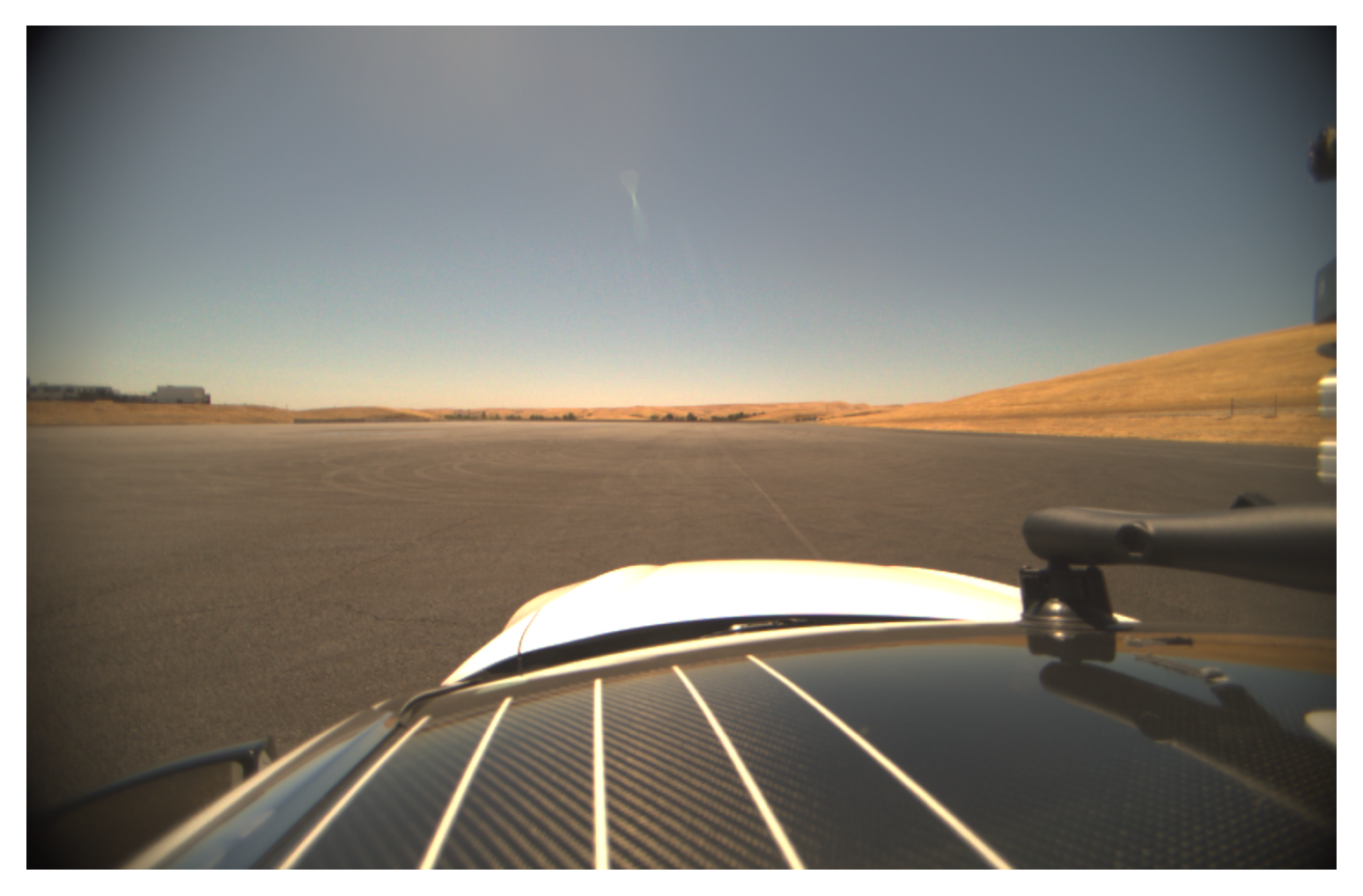}};
        \node[anchor=north west, fill=white, opacity=0.7, text opacity=1, font=\bfseries, xshift=5pt, yshift=-5pt] at (aug.north west) {August};
        \hfill
        
        \node[inner sep=0pt, right=1em of aug] (sep) {\includegraphics[width=0.47\columnwidth]{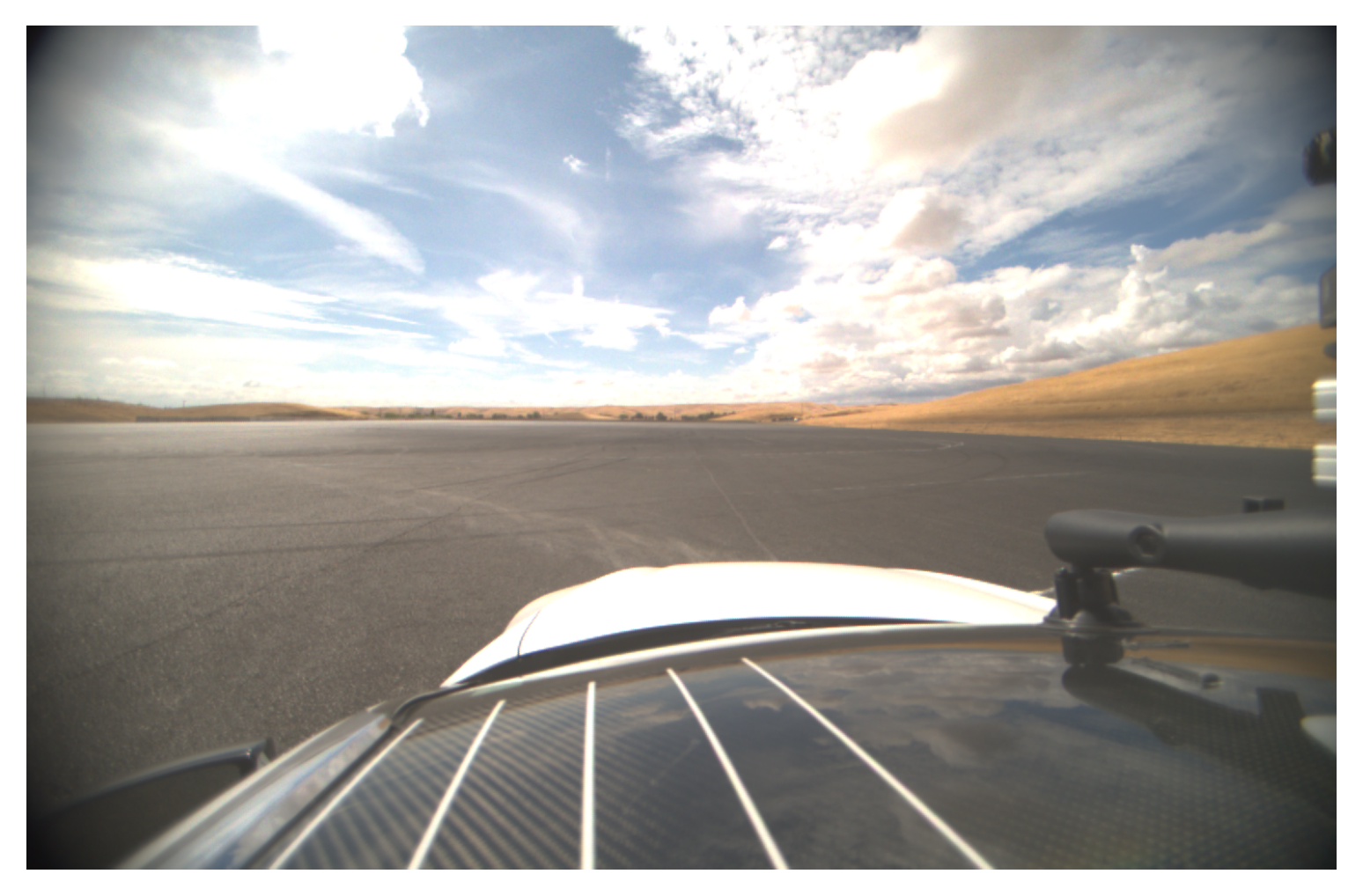}};
        \node[anchor=north west, fill=white, opacity=0.7, text opacity=1, font=\bfseries, xshift=5pt, yshift=-5pt] at (sep.north west) {September};
    \end{tikzpicture}%
    \vspace{-10pt}
    
    \caption{Difference in lighting conditions in the August and September tests.}
    \label{fig:lightingconditions}
\end{figure}

\begin{table}[t]
    \centering
    \caption{Percentage of successful attempted turns through a puddle.}
    \vspace{-5pt}
    \resizebox{\columnwidth}{!}{
    \begin{tabular}{ccccc}\toprule
    Occasion & \basemodelname & \basemodelname + RLS & \modelname ($\context_t = 0$) & \modelname \\\midrule
    August & 0 (0/1) & 0 (0/1) & -- & \textbf{100} (3/3)\\
    September & 0 (0/3) & 0 (0/1) & 0 (0/1)& \textbf{100} (9/9)\\\bottomrule
    \end{tabular}
    }
    \vspace{-1em}
    \label{tab:spinouts}
\end{table}

The hardware results were repeatable, as shown in Tab.~\ref{tab:spinouts}, with one set of experiments from the second row illustrated in Fig.~\ref{fig:hardware_res}. In total, across the two test sessions, the baselines spun out every single time (although the baseline \modelname ignoring vision was not tested in August).
In contrast, the \modelname with vision context completed 12 out of 12 attempted laps, indicating that it is robust to experimental conditions.

\section{Conclusion}
This letter presents a vision-conditioned VBLL vehicle model that enables proactive control in autonomous racing under varying surface conditions.
By conditioning the dynamics model on visual context via FiLM, a controller can anticipate changes in road surface properties before encountering them.
Our approach employs a two-stage training strategy: first learning a nominal VBLL model on dry surface data, then fine-tuning the conditioning path using the limited available data of water interactions.
Hardware experiments on a Lexus LC500 racing through water puddles demonstrate the efficacy of the approach: the vision-conditioned model completed multiple laps across varying conditions, while baseline methods without visual context consistently spun out.
These results demonstrate that incorporating exteroceptive information into learned dynamics models is essential for safe and robust autonomous racing in changing conditions. Future work will explore the use of vision-conditioned models for predictive control in other applications, such as autonomous drifting and off-road driving with a richer visual context.


\appendix

\fakepar{Physics-based vehicle model}\label{eq:bike:fiala}
The dynamics in \eqref{eq:learned_dynamics} are
\begin{subequations}\label{eq:bikedynamics}
\begin{align}
\dot{r} =& (
a
	F_{yf}\cos(\delta)
+
	aF_{xf}\sin(\delta)
 - 
b F_{yr})/I_z,
\\
\dot{v} =& \big(-F_{yf}\sin(\delta-\beta)+F_{xf}\cos(\delta-\beta)\notag
\\
&+ F_{yr}\sin\beta
+
F_{xr}\cos\beta
\big)/m,
\\
\dot{\beta} =& -r +
\big(
F_{yf}\cos(\delta-\beta)+F_{xf}\sin(\delta-\beta)\notag\\
&+ F_{yr}\cos\beta-F_{xr}\sin\beta\big)/(mv),
\\
\dot{\omega}_r =& (\tau_{\textrm{eng}}
-F_{xr}r_\textrm{w})/I_\textrm{w},
\end{align}
\end{subequations}
\begin{wraptable}{r}{2.5cm}
    \centering
    \vspace*{-15pt}
    \caption{Model param.}
    \vspace{-7pt}
    \resizebox{2.5cm}{!}{%
    \begin{tabular}{cc}\toprule
        Param. & Value \\\midrule
        $a$ & 1.3457754\\
        $b$ & 1.5222246\\
        $I_z$ & 3675\\
        $m$ & 2048\\
        $r_{\mathrm{w}}$ & 0.368\\
        $I_{\mathrm{w}}$ & 30 \\
        $C_f$ & 156000\\
        $C_r$ & 422000 \\
        $\mu_f$ & 1.02 \\
        $\mu_r$ & 1.08 \\\bottomrule
    \end{tabular}}
    \vspace{-10pt}
    \label{tab:parameters}
\end{wraptable}
with parameters $(a, b, I_z, m, r_{\mathrm{w}}, I_{\mathrm{w}})\in\Real_{>0}^6$. The tire forces $F_{xf},F_{yf},F_{xr},F_{yr}$ (N) are nonlinear functions of the state (see, e.g.~\cite{lew2025risk} for details) parametrized in cornering stiffness $(C_f, C_r)\in\Real_{>0}^2$ and friction coefficients $(\mu_f, \mu_r)\in\Real_{>0}^2$, with $f$ and $r$ referring to the front and rear wheels, respectively. The parameters are found by physical measurements and gray-box system identification. The parameters used in the experiments are summarized in SI units in Tab.~\ref{tab:parameters}.

\fakepar{Baseline VBLL Model and Training Details.}\label{app:trainingparameters}
For the backbone of the VBLL model, we use 4 hidden layers with 64 neurons and ELU activations.
The LSTM of the FiLM VBLL has a latent state size of 16 and we use an additional linear layer to decode this latent to the FiLM vectors (\cf Fig.~\ref{fig:overview}).
For the VBLL heads, we set $\xi_i^{\circ}=1$, $\alpha_i^{\circ}=\frac{1}{2}$, and $\beta_i^{\circ}=\frac{0.01}{2}$.
We implicitly set $\gamma^{\circ}$ by adding weight decay of $10^{-4}$ on the backbone as in \cite{harrison2024variationalbayesianlayers} and using AdamW \cite{loshchilov2018decoupled} with a learning rate of $5 \cdot 10^{-4}$ for the base model and $10^{-4}$ for fine-tuning.
We further use a gradient clipping of 1 for both.
For the base model, we train for \num{5000} epochs and for the fine-tuning for \num{1000}, but use the best performing model based on a holdout validation set.
For both, we use a batch size of \num{1028}.
The regularization parameters are $\lambda_{\text{reg}}=10$ and $\lambda_{\text{L2-SP}}=10$.

\fakepar{Controller parameters.}\label{app:gains}
The nominal MPC weights are $\lambda_T = 500$,
$\Q = \mathrm{diag}(10, 0.1, 1, 10^{-8}, 0.01, 10, 0.01, 0)$,
${\R\! =\! \mathrm{diag}(0.1, 10^{-4}, 10^{-5})}$,
${\T\! =\! \mathrm{diag}(2\!\cdot\! 10^{5}, 10^{-5}, 10^{-9})}$.

\section*{ACKNOWLEDGMENT}
The authors acknowledge Steven Goldine for driving in the data collection, Emre Adabag for helpful discussions, and William Kettle, Phung Nguyen, Zachary Conybeare, and Max Freeman for support in the experiments.

\bibliography{main.bib}


\end{document}